\documentclass[10pt,twocolumn,letterpaper]{article}
\usepackage[dvipsnames,table]{xcolor}
\usepackage{cvpr}

\definecolor{cvprblue}{rgb}{0.21,0.49,0.74}
\usepackage[pagebackref,breaklinks,colorlinks,allcolors=cvprblue]{hyperref}

\usepackage{amsmath}
\usepackage{amsfonts}
\usepackage{pifont}  
\usepackage{tikz}
\usepackage{pgfplots}
\usetikzlibrary{patterns}
\usepackage{caption}
\usepackage{graphicx}
\usepackage{tabularx}
\usetikzlibrary{arrows.meta}
\usetikzlibrary{shapes,positioning,intersections,quotes}

\usepackage[linesnumbered,ruled,vlined]{algorithm2e}

\SetCommentSty{mycommfont}

\usepackage{makecell}
\usepackage{multirow}
\usepackage{dsfont}

\newcommand{\PreserveBackslash}[1]{\let\temp=\\#1\let\\=\temp}
\newcolumntype{C}[1]{>{\PreserveBackslash\centering}p{#1}}
\newcolumntype{R}[1]{>{\PreserveBackslash\raggedleft}p{#1}}
\newcolumntype{L}[1]{>{\PreserveBackslash\raggedright}p{#1}}

\newcommand{\plugin}[1]{{\small #1}}
\newcommand{\gain}[1]{\color{red}{+\small #1}}

\usepgfplotslibrary{fillbetween}

\def\a{{\boldsymbol a}}

\def\p{{\boldsymbol p}}

\def\s{{\boldsymbol s}}

\def\x{{\boldsymbol x}}
\def\Y{{\boldsymbol Y}}
\def\y{{\boldsymbol y}}

\def\z{{\boldsymbol z}}

\def\BM{{\mathcal B}}

\def\DM{{\mathcal D}}

\def\HM{{\mathcal H}}

\def\NM{{\mathcal N}}

\def\PM{{\mathcal P}}

\def\RM{{\mathcal R}}

\def\XM{{\mathcal X}}

\def\EB{{\mathbb E}}

\def\softmax{\mathtt{softmax}}

\def\round{\mathtt{round}}

\def\hone{{\mathds{1}}}

\def\argmin{\mathop{\rm argmin}}

\def\method{{DUS}}
\def\CREMAD{{CREMA-D}}
\def\Kinetics{{Kinetics-Sounds}}
\def\Twitter{{Twitter2015}}
\def\Sarcasm{{Sarcasm}}
\def\NVGesture{{NVGesture}}
\newcommand\unimodal[1]{\cellcolor{gray!25}#1}
\newcommand\first[1]{\bf{#1}}
\newcommand\second[1]{\underline{#1}}

\def\paperID{705} 
\def\confName{CVPR}
\def\confYear{2025}

\title{Rebalanced Multimodal Learning with Data-aware Unimodal Sampling}

\author{
Qingyuan Jiang$^1$ \quad
Zhouyang Chi$^1$ \quad
Xiao Ma$^2$ \quad
Qirong Mao$^3$ \quad
Yang Yang$^1$ \quad
Jinhui Tang$^1$
\\
\small{$^1$ School of Computer Science and Engineering, Nanjing University of Science and Technology} \\
\small{$^2$ School of Computer Science and Technology, Zhejiang Sci-Tech University} \\
\small{$^3$ School of Computer Science and Communication Engineering, Jiangsu University}\\
\small{$^1$\texttt{\{jiangqy,122106222804,yyang,jinhuitang\}@njust.edu.cn}}\\
\small{$^2$\texttt{mxsujin94@gmail.com}}\\
\small{$^3$\texttt{mao\_qr@ujs.edu.cn}}
}
\begin{document}
\maketitle
\begin{abstract}
To address the modality learning degeneration caused by modality imbalance, existing multimodal learning~(MML) approaches primarily attempt to balance the optimization process of each modality from the perspective of model learning. However, almost all existing methods ignore the modality imbalance caused by unimodal data sampling, i.e., equal unimodal data sampling often results in discrepancies in informational content, leading to modality imbalance. Therefore, in this paper, we propose a novel MML approach called \underline{D}ata-aware \underline{U}nimodal \underline{S}ampling~(\method), which aims to dynamically alleviate the modality imbalance caused by sampling. Specifically, we first propose a novel cumulative modality discrepancy to monitor the multimodal learning process. Based on the learning status, we propose a heuristic and a reinforcement learning~(RL)-based data-aware unimodal sampling approaches to adaptively determine the quantity of sampled data at each iteration, thus alleviating the modality imbalance from the perspective of sampling. Meanwhile, our method can be seamlessly incorporated into almost all existing multimodal learning approaches as a plugin. Experiments demonstrate that \method~can achieve the best performance by comparing with diverse state-of-the-art~(SOTA) baselines.
\end{abstract}

\section{Introduction}
Multimodal learning~\cite{MML:conf/ijcai/YangWZX019,MML:journals/pami/BaltrusaitisAM19,MML:journals/corr/abs-2301-04856} has become an active research topic in recent years. The goal of MML is to develop robust representations of multimodal data to improve performance in different application scenarios~\cite{AVSR:journals/cm/YuhasGS89,Classification:journals/mta/LanBYLH14,CMR:journals/corr/WangY0W016} including speech recognition~\cite{AVSR:journals/cm/YuhasGS89,MMDL:conf/icml/NgiamKKNLN11}, action classification~\cite{Classification:journals/mta/LanBYLH14}, information retrieval~\cite{CMR:journals/corr/WangY0W016,CBGL:journals/tmm/ZhangLL24}, and so on.

In real-world scenarios, multimodal data collected from different sensors exhibits significant heterogeneity. Due to the heterogeneity, recent research~\cite{OGR-GB:conf/cvpr/WangTF20} has identified a counterintuitive phenomenon where MML performs worse than unimodal models under specific conditions. Essentially, this is due to the existence of dominant and non-dominant modalities. These individual modalities will converge at different rate~\cite{OGM:conf/cvpr/PengWD0H22}, thus affecting the performance.

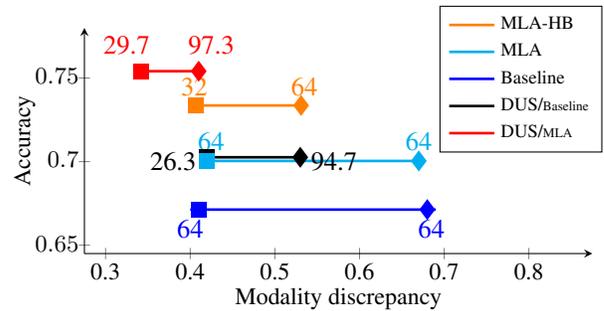
\begin{figure}[t]
\centering
\begin{tikzpicture}
\begin{axis}[
legend cell align={left},
inner sep=0pt,
outer sep=0pt,
axis x line=center,
axis y line=center,
height={0.55\linewidth},
width={\linewidth},
xtick={0.3,0.4,0.5,0.6,0.7,0.8},
typeset ticklabels with strut,
enlarge x limits=false,
xticklabels={0.3,0.4,0.5,0.6,0.7,0.8},
ytick={0, 0.05,...,0.8},
xmin=0.275,
xmax=0.875,
ymin=0.645,
ymax=0.78,
xlabel=Modality discrepancy,
ylabel=Accuracy,
xlabel near ticks,
ylabel near ticks,
xticklabel style = {font=\small},
xlabel style = {font=\small},
yticklabel style = {font=\small},
ylabel style = {font=\small},
legend style={at={(0.7,0.45)},anchor=south west},font=\scriptsize ]
\addplot [line width=0.35mm,color=orange] table {
0.407  0.7336
0.531  0.7336
};\addlegendentry{MLA-HB}
\addplot [line width=0.35mm,color=cyan] table {
0.42  0.7004
0.67  0.7004
};\addlegendentry{MLA}
\addplot [line width=0.35mm,color=blue] table {
0.4   0.6713
0.69  0.6713
};\addlegendentry{Baseline}
\addplot [line width=0.35mm,color=black] table {
0.42  0.7026
0.53  0.7026
};\addlegendentry{\method/{\tiny Baseline}}
\addplot [line width=0.35mm,color=red] table {
0.342  0.7541
0.41  0.7541
};\addlegendentry{\method/{\tiny MLA}}
\addplot [draw opacity=0, only marks, mark=square*,mark size=3pt,fill=blue] table {
0.41  0.6713
};
\addplot [draw opacity=0, only marks, mark=diamond*,mark size=4pt,fill=blue] table {
0.68  0.6713
};
\addplot [draw opacity=0, only marks, mark=square*,mark size=3pt,fill=black] table {
0.42  0.7026
};
\addplot [draw opacity=0, only marks, mark=diamond*,mark size=4pt,fill=black] table {
0.53  0.7026
};
\addplot [draw opacity=0, only marks, mark=square*,mark size=3pt,fill=cyan] table {
0.42  0.7004
};
\addplot [draw opacity=0, only marks, mark=diamond*,mark size=4pt,fill=cyan] table {
0.67  0.7004
};
\addplot [draw opacity=0, only marks, mark=square*,mark size=3pt,fill=orange] table {
0.407  0.7336
};
\addplot [draw opacity=0, only marks, mark=diamond*,mark size=4pt,fill=orange] table {
0.531  0.7336
};
\addplot [draw opacity=0, only marks, mark=square*,mark size=3pt,fill=red] table {
0.342  0.7541
};
\addplot [draw opacity=0, only marks, mark=diamond*,mark size=4pt,fill=red] table {
0.41  0.7541
};
\node[] at (axis cs: 0.325,0.77) {\normalsize {\color{red}29.7}};
\node[] at (axis cs: 0.425,0.77) {\normalsize {\color{red}97.3}};
\node[] at (axis cs: 0.425,0.712) {\normalsize {\color{cyan}64}};
\node[] at (axis cs: 0.67,0.712) {\normalsize {\color{cyan}64}};
\node[] at (axis cs: 0.38,0.70) {\normalsize {\color{black}26.3}};
\node[] at (axis cs: 0.57,0.70) {\normalsize {\color{black}94.7}};
\node[] at (axis cs: 0.40,0.66) {\normalsize {\color{blue}64}};
\node[] at (axis cs: 0.685,0.66) {\normalsize {\color{blue}64}};
\node[] at (axis cs: 0.405,0.745) {\normalsize {\color{orange}32}};
\node[] at (axis cs: 0.535,0.745) {\normalsize {\color{orange}64}};
\end{axis}
\end{tikzpicture}
\caption{Relationship between performance and the quantity of sampled data on \Kinetics~dataset, where the rectangle and diamond markers denote the video and audio modalities, respectively. The average batch size is marked with its corresponding colors around the markers. By adjusting the batch size, we can affect modality discrepancy and thereby improve modality learning.}
\label{fig:motivation}
\end{figure}

Many impressive works~\cite{OGR-GB:conf/cvpr/WangTF20,SMV:conf/cvpr/YakeRZD24,OGM:conf/cvpr/PengWD0H22,MLA:conf/cvpr/ZhangYBY24,MAIE:journals/corr/abs-2407-04587,DI-MML:conf/mm/FanXWLG24,ReconBoost:conf/icml/CongHua24} have been proposed to rebalance the multimodal learning in recent years. Compared with general multimodal learning, these approaches typically establish connections between the training processes of individual modalities. Some of these methods leverage key information from the modality training process, such as gradients~\cite{OGR-GB:conf/cvpr/WangTF20,OGM:conf/cvpr/PengWD0H22,AGM:conf/iccv/LiLHLLZ23} and learning rates~\cite{MSLR:conf/acl/YaoM22}, to balance the learning of different modalities by utilizing these information to adjust the fitting speed of dominant and non-dominant modalities. Other attempts~\cite{MLA:conf/cvpr/ZhangYBY24,DI-MML:conf/mm/FanXWLG24,ReconBoost:conf/icml/CongHua24,MAIE:journals/corr/abs-2407-04587} explore the training paradigms for multimodal learning, and design an alternating learning strategy to learn multimodal models. In summary, these methods mitigate modality imbalance to some extent by balancing the learning across different modalities. 

However, almost all existing methods primarily address modality imbalance from the model learning perspective while ignoring another key factor, unimodal data sampling. Due to the heterogeneity of data, the same quantity of data often contains different information content. That is to say, the dominant and non-dominant modalities will provide different information content for training at each iteration if the model is trained with the same quantity of data. Models trained with different information content will still encounter the problem of modality imbalance caused by the equal quantity of sampled data, which will ultimately affect performance. To support this viewpoint, we conduct a toy experiment to explore the relationship between the quantity of sampled data~(\#batch size) and overall performance. We utilize the modality discrepancy score defined by on-the-fly gradient modulation~(OGM)~\cite{OGM:conf/cvpr/PengWD0H22} to bridge their relationship more precisely. The modality discrepancy refers to the average prediction confidence of the correct class. We compare the vanilla MML approach~(Baseline) which minimizes the unimodal and multimodal losses, a competitive MML approach MLA~\cite{MLA:conf/cvpr/ZhangYBY24}, and a variant of MLA~(MLA-HB) in Figure~\ref{fig:motivation}, where the rectangle and diamond markers respectively denote the video and audio modalities, and the average batch size is marked with its corresponding colors around the markers. The MLA-HB denotes that we set the batch size of dominant modality~(audio) to half of the non-dominant modality~(video), i.e., 32 for audio and 64 for video. An interesting phenomenon can be observed: equal quantities of multimodal data often lead to a larger modality discrepancy gap. By slightly reducing the amount of data from the dominant modality, this discrepancy can be reduced, thereby improving overall performance.

Based on our findings, we can control the information content each modality provides during training by adjusting the quantity of sampled data. More precisely, we first modify the modality discrepancy score by cumulatively averaging the modal's predictions of the ground-truth class for data points at each iteration. In this way, we can dynamically capture the learning status of each modality during training, thereby guiding data sampling. Then, we propose a heuristic and a reinforcement learning-based adaptive unimodal sampling approaches. The former approach employs a heuristic strategy to reduce the quantity of dominant modality, thus rebalancing the learning of each modality. Meanwhile, we propose another adaptive unimodal sampling approach by using reinforcement learning. In Figure~\ref{fig:motivation}, we also illustrate the results of RL-based \method~methods which have been integrated with baseline~(\method/\plugin{Baseline}) and MLA~(\method/\plugin{MLA}). By adaptive unimodal sampling, \method~achieves the smallest modality discrepancy gap and the best accuracy with lower batch size for audio and higher batch size for video, demonstrating the necessity of dynamically adjusting the quantity of sampling. Furthermore, there also appears one approach SMV~\cite{SMV:conf/cvpr/YakeRZD24} which addresses modality imbalance from the data sampling perspective. SMV focuses on re-sampling data points with a lower contribution. To sum up, our contributions are listed as follows:
\begin{itemize}
\item By averaging the model's predictions of the ground-truth class, we design a cumulative modality discrepancy score to monitor the learning status for interactive MML.
\item Based on the discrepancy score, we propose a heuristic and an RL-based adaptive unimodal sampling approaches to dynamically adjust the quantity of unimodal sampled data. Meanwhile, our method can be utilized as a plug-and-play for various interactive MML methods.
\item Extensive experiments demonstrate that \method~can achieve the best performance by comparing with various SOTA baselines across widely used datasets.
\end{itemize}

\section{Related Work}\label{sec:related-work}
\subsection{Rebalanced Multimodal Learning}
In multimodal learning, recent works~\cite{OGM:conf/cvpr/PengWD0H22} have revealed a counterintuitive phenomenon where MML performs worse than unimodal models. The reason behind this phenomenon is modality imbalance~\cite{OGM:conf/cvpr/PengWD0H22,ModalCompetition:conf/icml/HuangLZYH22}. Due to the existence of dominant and non-dominant modalities, the individual modalities will converge at different speeds~\cite {OGM:conf/cvpr/PengWD0H22}. 

Naturally, some researchers have proposed a series of methods~\cite{OGR-GB:conf/cvpr/WangTF20,OGM:conf/cvpr/PengWD0H22,PMR:conf/cvpr/Fan0WW023,AGM:conf/iccv/LiLHLLZ23,MMPareto:conf/icml/WeiH24} to solve the problem of modality imbalance through balancing modality learning. To be more specific, these methods attempt to slow down the learning of the strong modality by adjusting the gradients to ensure that the learning of both modalities is as balanced as possible. Other attempts~\cite{Greedy:conf/icml/WuJCG22,UMT:conf/icml/DuTLLYWYZ23,CBGL:journals/tmm/ZhangLL24} try to introduce some extra modules as auxiliary to rebalance the modality learning. Given that multimodal training is relatively independent, recent works~\cite{MLA:conf/cvpr/ZhangYBY24,DI-MML:conf/mm/FanXWLG24,MAIE:journals/corr/abs-2407-04587} have sought to enhance interactions between modalities to balance the learning process. By leveraging gradients~\cite{MLA:conf/cvpr/ZhangYBY24,MAIE:journals/corr/abs-2407-04587} or deep features~\cite{DI-MML:conf/mm/FanXWLG24}, these methods can assess the learning status of the modalities during the process, thereby rebalancing the modality learning to some extent. However, these methods overlook the modality imbalance caused by modality discrepancy derived from the same quantity of sampled data.

\begin{figure*}
\centering
 \includegraphics[width=0.875\linewidth]{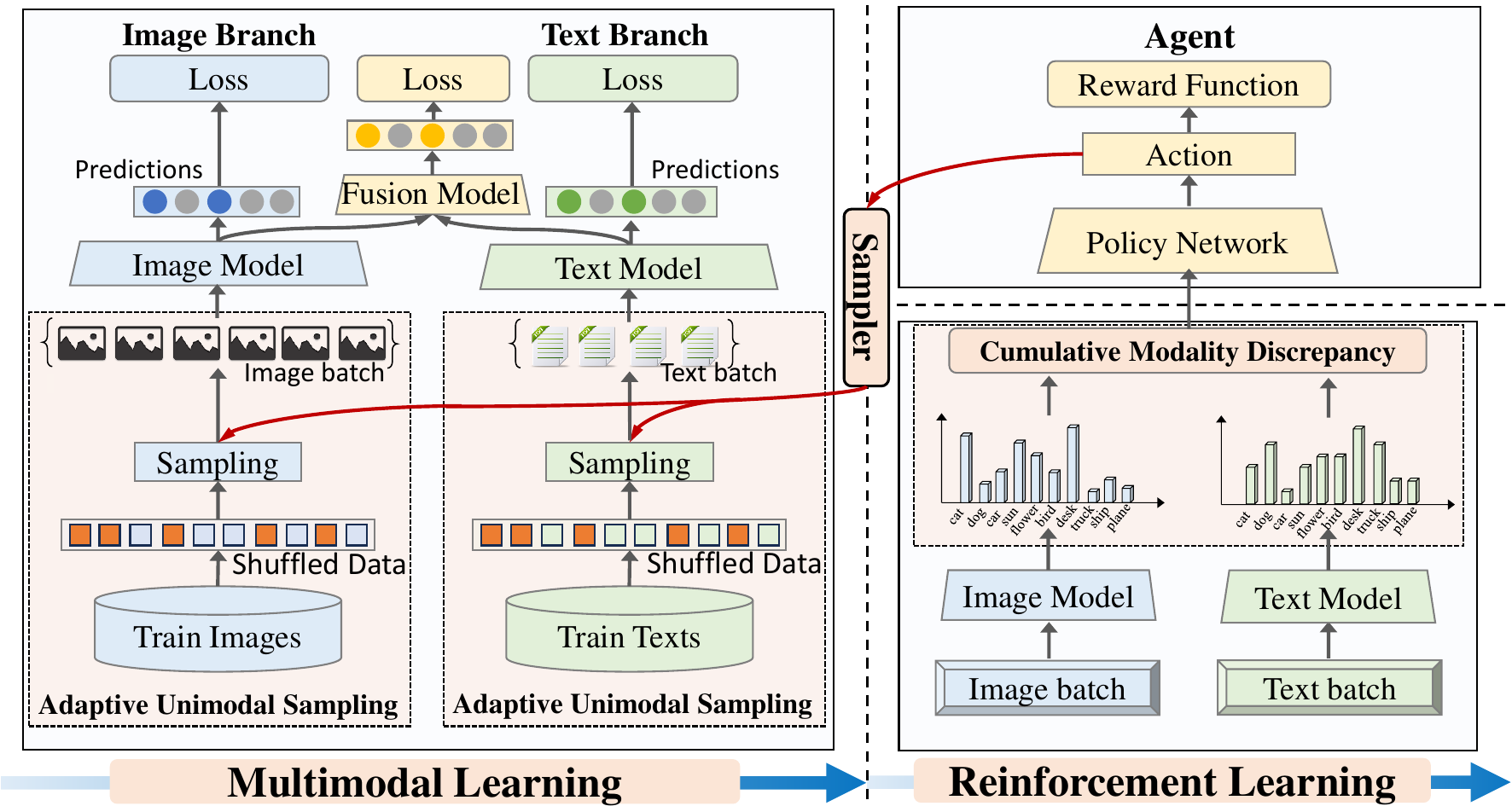}
\caption{The architecture of our proposed \method~using the RL-based adaptive unimodal sampling as an example. \method~contains two important components, i.e., cumulative modality discrepancy calculation and adaptive unimodal sampling.}
\label{fig:framework}
\end{figure*}

\subsection{Reinforcement Learning}

Reinforcement learning has achieved much progress in various domains, such as large language models~\cite{RLHF:conf/nips/Ouyang0JAWMZASR22,GPT-3:conf/nips/BrownETC20}, game playing~\cite{RL:journals/nature/SilverHMGSDSAPL16}, robotics~\cite{IsaacGym:conf/nips/MakoviychukWGLS21}, and so on. 
In reinforcement learning, an agent learns an effective policy by maximizing returns from trial-and-error interactions with the environment.
Based on this policy, the agent can take good actions and receive high rewards from the environment.
Hence, we adopt reinforcement learning to solve the problem of how to determine the amount of training data, which is detailedly described in the following section.
One of the classic and easy-to-implement reinforcement learning methods is REINFORCE~\cite{Reinforcement:journals/ml/Williams92}.
The goal of REINFORCE is to maximize the expected returns by adjusting the policy parameters. It does this by calculating the gradient of the expected reward with respect to the policy parameters.

\section{Methodology}\label{sec:method}
The architecture of \method~is illustrated in Figure~\ref{fig:framework}. We first provide the calculation of the cumulative modal discrepancy. Then, we illustrate the heuristic and the reinforcement learning-based adaptive unimodal sampling methods to adaptively determine the quantity of sampled data. 

\subsection{Preliminary}

Without loss of generality, we suppose that a multimodal data point is defined as $\x=\{\x^{(1)},\cdots,\x^{(m)}\}$, where $m$ is the number of modalities. The category label is also available and defined as $\y\in\{0,1\}^c$, where $c$ denotes the number of category labels. In multimodal learning, we are interested in seeking a hypothesis $h(\cdot)$ that describes the relationship between the multimodal data $\x$ and the corresponding labels $\y$, which follow the joint distribution $\PM$. We usually adopt a loss function $\ell$ to penalize the differences between predictions $h(\x)$ and labels $\y$. Our goal can be formed as the following optimization problem:
\begin{align}
h^*=\argmin_{h\in\HM}\;R(h)=\EB_{(\x,\y)\sim\PM}[\ell(h(\x),h(\y))],\label{pro:er}
\end{align}
where $\EB(\cdot)$ denotes the expectation and $\HM$ denotes the hypothesis set. In practice, as the distribution $\PM$ is usually unknown, we try to optimize the empirical risk minimization~(ERM) over an observed dataset $\DM=\{\x_1,\cdots,\x_n\}$:
\begin{align}
\min\;L(h)=\frac{1}{n}\sum_{i=1}^n\ell(h(\x_i),\y_i).\nonumber
\end{align}

The discrepancy of dataset $\DM$ determines how well the ERM can approximate the optima of the problem~(\ref{pro:er}). In general, we can affect model performance by adjusting the data quantity as it influences the dataset discrepancy.

Then, we provide some details about how to construct the hypothesis $h(\cdot)$. Following representative MML methods~\cite{OGM:conf/cvpr/PengWD0H22,AGM:conf/iccv/LiLHLLZ23}, we also adopt deep neural networks~(DNN) as basic encoders. Specifically, we use $\phi^{(j)}(\cdot)$ to denote the encoder of $j$-th modality. After that, we employ a classification head $g^{(j)}(\cdot)$ to map the feature into $\RM^c$ space, which can be formed as follows:
\begin{align}
\z^{(j)}=g^{(j)}(\phi^{(j)}(\x^{(j)})).\nonumber
\end{align}
Once we obtain the $\z^{(j)}$, we further utilize a softmax layer to generate the prediction of $j$-th modality:
\begin{align}
\p^{(j)}=\softmax(\z^{(j)}).\nonumber
\end{align}
Taking cross-entropy loss, the ERM problem can be presented as:
\begin{align}
\min\;L(\XM^{(j)},\Y)
&=-\frac{1}{n}\sum_{i=1}^n\y_i^\top\log(\p_i^{(j)}).\label{eq:obj}
\end{align}

Multimodal learning models are usually trained through optimizing Problem~(\ref{eq:obj}). To address modality imbalance issue, existing MML methods~\cite{OGR-GB:conf/cvpr/WangTF20,OGM:conf/cvpr/PengWD0H22,Greedy:conf/icml/WuJCG22,AGM:conf/iccv/LiLHLLZ23,UMT:conf/icml/DuTLLYWYZ23,MLA:conf/cvpr/ZhangYBY24,ReconBoost:conf/icml/CongHua24,DI-MML:conf/mm/FanXWLG24} primarily attempt to balance the optimization process of each modality by adjusting the learning process using the gradient, learning rate, or effective features. In other words, these methods try to rebalance the multimodal learning from the model learning perspective.

\subsection{Cumulative Modality Discrepancy Score}\label{sec:modality-discrepancy}
Inspired by OGM~\cite{OGM:conf/cvpr/PengWD0H22}, we first employ the discrepancy score~\cite{OGM:conf/cvpr/PengWD0H22} to evaluate the discrepancy of different modalities during the learning process. Specifically, assuming that at $t$-th iteration, we are given a batch data points $\BM^{(j)}=\{\x^{(j)}_{1},\x^{(j)}_{2},\cdots,\x^{(j)}_{n_b}\}$. We first calculate their features and predictions by:
\begin{align}
\forall i\in\{1,\cdots,n_b\},\;&\z^{(j)}_i=g^{(j)}(\phi^{(j)}(\x^{(j)}_i)),\nonumber\\
&\p^{(j)}_i=\softmax(\z^{(j)}_i).\nonumber
\end{align}
Then the discrepancy score is defined as:
\begin{align}
s^{(j)}_t&=\frac{1}{n_b}\sum_{i=1}^{n_b}\y_i^\top\softmax(\z^{(j)}_i)=\frac{1}{n_b}\sum_{i=1}^{n_b}\y_i^\top\p^{(j)}_i,\nonumber
\end{align}
Due to the randomness of a single batch, we further define a cumulative discrepancy score based on discrepancy score~\cite{OGM:conf/cvpr/PengWD0H22} by:
\begin{align}
\left\{
\begin{aligned}
&\hat s^{(j)}_t=s^{(j)}_t,\;&\text{if }t=1,\\
&\hat s^{(j)}_t=\frac{t-1}{t}\hat s^{(j)}_{t-1}+\frac{1}{t}s^{(j)}_t,\;&\text{otherwise.}\label{eq:discrepancy}
\end{aligned}
\right.
\end{align}

In Equation~(\ref{eq:discrepancy}), $\hat s^{(j)}_t$ is defined as the average confidence score of the ground-truth class of all samples cumulatively. Hence, $\hat s^{(j)}_t$ characterizes the discrepancy from two perspectives: the number of samples in current batch, i.e., batch size, and model output. When the model has enough ability to distinguish the data, i.e., a higher discrepancy score, its training should be appropriately suppressed. On the contrary, when the prediction scores given by the model for the ground-truth category is relatively low, i.e., a lower discrepancy score, we should enhance the training of the corresponding model.

\subsection{Adaptive Unimodal Sampling}

Generally, we observe that the discrepancy of the dominant modality is higher at the beginning of training. Therefore, we should reduce the influence of dominant modalities at the start. Gradually, we can recover the learning of the dominant modality, allowing the model to learn multimodal information in a balanced manner. Hence, we design a heuristic adaptive unimodal sampling strategy to achieve this goal.

Specifically, we utilize the following formula to calculate the quantity of data points, i.e., batch size, at each epoch $T$:
\begin{align}
f(T)=\round(\beta e^{\alpha T}\times N_B),\nonumber
\end{align}
where $\round(\cdot)$ denotes the rounding function, and $N_B$ is a constant used to denote the initial batch size. $\alpha>0,\beta > 0$ are parameters used to guarantee that: (1). The batch size for dominant modality is smaller than that of non-dominant modality at the beginning of training; (2). At the end of the training, the batch size for the dominant modality is equal to the batch size of the non-dominant modality. The specific values of $\alpha$ and $\beta$ will be discussed in the experiment part.

To better characterize the changes in discrepancy during the learning process to balance the learning process, we further define the problem of how to determine the quantity of training data for each iteration based on discrepancy evaluation metrics as a reinforcement learning problem.

Specifically, we define the multimodal training phase as the \textit{Environment} in reinforcement learning. The modality-oriented discrepancy scores supplied by the environment construct the \textit{State Space} $\mathcal{S}$. At $t$-th iteration, the cumulative modality discrepancy scores $\{\hat s^{(j)}_t\}_{j=1}^m$ form a state vector $\hat\s_t$ as follows:
\begin{align}
\hat\s_t=[\hat s^{(1)}_t,\cdots,\hat s^{(m)}_t]^\top\in \mathcal{S}.\nonumber
\end{align}
Then, the \textit{Action Space} is defined as $\mathcal{A}  \circeq \NM_+^m$, where $\NM_+^m$ denotes the $m$-dimension positive integers.
At the $t$-th iteration, the action vector $\a_t$ is defined as follows:
\begin{align}
\a_t=[a^{(1)}_t,\cdots,a^{(m)}_t]^\top\in \mathcal{A},\nonumber
\end{align}
where $a_{t}^{(j)}$ represents the number of data points to be randomly sampled for the $j$-th modality from the shuffled dataset.

\begin{algorithm}[t]
\SetKwInOut{Input}{Input}
\SetKwInOut{Output}{Output}
\SetKw{KwBy}{by}
\SetInd{0.1em}{0.75em}
\DontPrintSemicolon
\SetAlgoLined
\SetNoFillComment
\Input{Training set $\DM$ and labels $\Y$.}
\Output{The learned parameters $\{\theta^{(j)}\}_{j=1}^{m}$.}
\textbf{INIT} initialize action $\a_1=[N_B/m,\cdots, N_B/m]^\top$\;
\For{$T=1\rightarrow \text{Epochs}$}{
    \tcc{\bf Learn models based on vanilla MML.}
    \For{$i=1\rightarrow \text{Num\_batch}$}{
        \For{$j=1\rightarrow m$}{
            Sample a mini-batch $\BM^{(j)}$ based on $a_{t}^{(j)}$.\;
            Calculate the feature $\z^{(j)}_i$ for $\x^{(j)}_i\in\BM^{(j)}$.\;
            Calculate loss function according to Eq.~(\ref{eq:obj}).\;
            Calculate the gradient $\nabla_{\theta^{(j)}}L$.\;
            Update the parameters according to gradient.\;
        }
    }
    \tcc{\bf Learn batch size based on RL.}
    Update action vector $\a_t$. \;
    Receive a state $\s_t$ according to the discrepancy score.\;
    Choose an action based on the policy and Eq.~(\ref{eq:a_{t}^{(j)}}).\;
    Receive a reward based on Eq.~(\ref{eq:reward_function}).\;
    Update the policy network based on Eq.~(\ref{eq:gradient-RL}).\;
    Update $t$: $t=t+1$.\;
}
\caption{The Learning Algorithm for \method.}
\label{algo:ours}
\end{algorithm}

Given a state vector, our goal is to learn a \textit{Policy Network} $\psi_{\omega}$ to generate an action vector, where $\omega$ is the parameter of the policy network.
Please note that the action vector consists of positive integers.
Due to the difficulty of generating discrete values, for the convenience of training, we use $\hat\a_t\in [0,1]^m$ as the output of the policy network, shown as follows:
\begin{align}
\hat\a_t=\psi_{\omega}(\hat\s_t),\nonumber
\end{align}
where the action $\hat{a}_{t}^{(j)}$ in $\hat\a_t$ indicates the proportion of data for $j$-th modality.
To establish the connection between the output of the policy network and the action, we assume the total number of data for all modalities is a constant $N_B$ and the action $a_{t}^{(j)}$ for $j$-th modality can be calculated by:
\begin{align}
a_{t}^{(j)}=\round(N_B\times\hat a_{t}^{(j)}), 
\label{eq:a_{t}^{(j)}}
\end{align}
Furthermore, \textit{Reward Function} $r(\hat\s_t, \a_t, \hat\a_t)$ is defined as follows:
\begin{align}
r(\hat\s_t, \a_t, \hat\a_t)=-\frac{1}{m}\sum_{j=1}^m\hone({\hat s_{t}^{(j)}\neq\max{\{\hat s_{t}^{(k)}\}_{k=1}^m}})\log(\hat a_{t}^{(j)}),
\label{eq:reward_function}
\end{align}
where $\hone(\cdot)$ denotes the indicator function, i.e., $\hone(true)=1$ and $\hone(false)=0$. Then, we employ a classic reinforcement learning algorithm, REINFORCE~\cite{Reinforcement:journals/ml/Williams92}
to maximum the training objective $\EB_{\psi_{\omega}}[r(\hat\s_t, \a_t, \hat\a_t)]$ during $t$-th iteration.
Specifically, we obtain the gradient of $\omega$,
\begin{align}
\nabla_{\omega}\EB_{\psi_{\omega}}&[r(\hat\s_t, \a_t, \hat\a_t)]\nonumber\\&
=\EB_{\psi_{\omega}}[r(\hat\s_t, \a_t, \hat\a_t)\nabla_\omega\log(\psi_\omega(\hat\s_t, \hat\a_t))],\label{eq:gradient-RL}
\end{align}
and then update the parameter of the policy network. 
After we obtain the action vector $\a_t$ at $t$-th iteration, we randomly sample $a_{t}^{(j)}$ data points for $j$-th modality. The learning algorithm of \method~is summarized in Algorithm~\ref{algo:ours}. In Algorithm~\ref{algo:ours}, we utilize the vanilla MML algorithm as an example. One can substitute other MML algorithms to the vanilla MML approach and then integrate the \method~with them.
  
\noindent{\bf Discussion:} Our proposed method focuses on rebalancing multimodal learning from the data sampling perspective. Thus, our \method~requires only adjustments to the data sampling module and can be integrated with nearly all existing methods in a plug-and-play manner.

\section{Experiments}\label{sec:exp}
\subsection{Dataset}
We select five widely used datasets for evaluation. They are \Twitter~\cite{Twitter15:conf/ijcai/Yu019}, \Sarcasm~\cite{Sarcasm:conf/acl/CaiCW19}, \CREMAD~\cite{CREMAD:journals/taffco/CaoCKGNV14}, \Kinetics~\cite{Kinetics-Sound:conf/iccv/ArandjelovicZ17}, and \NVGesture~\cite{NVGeasture:conf/cvpr/MolchanovYGKTK16} datasets. The first two datasets, i.e., \Twitter~and \Sarcasm~datasets, contain image and text modalities and are collected from Twitter. These two datasets are used for emotion recognition and sarcasm detection task, respectively. \Twitter~dataset consists of 5,338 image-text pairs with 3,179 for training, 1,122 for validation, and 1,037 for testing. And \Sarcasm~dataset consists of 24,635 image-text pairs with 19,816 for training, 2,410 for validation, and 2,409 for testing. The \CREMAD~and \Kinetics~datasets contain audio and video modality. These two datasets are used for speech emotion recognition and video action recognition tasks, respectively. \CREMAD~dataset contains 7,442 2$\sim$3 second clips collected from 91 different actors. These clips are divided into 6,698 samples as the training set and 744 samples as the testing set. \Kinetics~dataset comprises 31 human action category labels and contains 19,000 10-second clips. \Kinetics~dataset is divided into a training set with 15K samples, a validation set with 1.9K samples, and a testing set with 1.9K samples. For the last dataset \NVGesture, we use RGB, Depth, and optical flow~(OF) modalities for experiments. This dataset contains 1,532 dynamic hand gestures and is divided into 1,050 for training and 482 for testing. It is worth mentioning that this dataset is used to verify our approach in scenarios with more than two modalities.

\begin{table*}[t]
\centering
\caption{Comparison with SOTA multimodal learning approaches, where the best and the second best are denoted as bold and underlining, respectively. The results with the gray background are based on MML but perform worse than the best unimodal approach.} 
\label{tab:main-exp}
\begin{tabular}{l|cc|cc|cc|cc}
\Xcline{1-9}{1pt}\hline
\multirow{2}{*}{Method}  & \multicolumn{2}{c|}{{\Twitter}}& \multicolumn{2}{c|}{{\Sarcasm}}& \multicolumn{2}{c|}{{\CREMAD}}& \multicolumn{2}{c}{{\Kinetics}}\\
\cline{2-9}
          & Acc.  & MacF1& Acc.  & MacF1& Acc.  & MAP& Acc.  & MAP\\
\hline\hline
Text/Video      & 73.67\% & 68.49\% & 81.36\% & 80.65\% & 63.17\% & 68.61\% & 53.12\% & 56.69\% \\
Image/Audio     & 58.63\% & 43.33\% & 71.81\% & 70.73\% & 45.83\% & 58.79\% & 54.62\% & 58.37\% \\
\hline
Baseline &73.94\% & 65.63\%& 82.46\%& 81.69\%& 68.87\%& 73.16\%& 67.13\%& 71.48\%\\
OGR-GB~\cite{OGR-GB:conf/cvpr/WangTF20}    & 74.35\% & 68.69\% & 83.35\% & 82.71\%& 64.65\% & \unimodal{68.54\%} & 67.10\% & 71.39\% \\
OGM~\cite{OGM:conf/cvpr/PengWD0H22}       & \second{74.92\%} & 68.74\% & 83.23\% & 82.66\% & 66.94\% & 71.73\% & 66.06\% & 71.44\% \\
DOMFN~\cite{DOMFN:conf/mm/0074ZGGZ22}     & 74.45\% & 68.57\% & 83.56\% & 82.62\% & 67.34\% & 73.72\% & 66.25\% & 72.44\% \\
MSES~\cite{MSES:conf/acpr/FujimoriEKM19}      & \unimodal{71.84\%} & \unimodal{66.55\%} & 84.18\% & 83.60\% & \unimodal{61.56\%} & \unimodal{66.83\%} & 64.71\% & 70.63\% \\
PMR~\cite{PMR:conf/cvpr/Fan0WW023}       & 74.25\% & 68.60\% & 83.60\% & 82.49\% & 66.59\% & 70.30\% & 66.56\% & 71.93\% \\
AGM~\cite{AGM:conf/iccv/LiLHLLZ23}       & 74.83\% & \first{69.11\%} & 84.02\% & 83.44\% & 67.07\% & 73.58\% & 66.02\% & 72.52\%  \\
MSLR~\cite{MSLR:conf/acl/YaoM22}      & \unimodal{72.52\%} & \unimodal{64.39\%} & 84.23\% & \second{83.69\%} & 65.46\% & 71.38\% & 65.91\% & 71.96\% \\
SMV~\cite{SMV:conf/cvpr/YakeRZD24}       & 74.28\% & 68.17\% & 84.18\% & 83.68\%& 78.72\% & 84.17\% & 69.00\% & 74.26\% \\
ReconBoost~\cite{ReconBoost:conf/icml/CongHua24}& 74.42\% & 68.34\% & 84.37\% & 83.17\% & 74.84\% & 81.24\% & 70.85\% & 74.24\% \\
DI-MML~\cite{DI-MML:conf/mm/FanXWLG24}    &  \unimodal{72.48\%}       &     \unimodal{66.86\%}    &    84.11\%     &    83.15\%    &   \second{81.58\%}       &   \second{85.92\%}     &     72.03\%    &  76.24\% \\
MLA~\cite{MLA:conf/cvpr/ZhangYBY24}       & \unimodal{73.52\%} & \unimodal{67.13\%} & 84.26\% & 83.48\% & 79.43\% & 85.72\% & 70.04\% & 74.13\% \\
\hline
\method/\plugin{Baseline} &74.32\% &68.22\% & 84.20\%& 83.76\%& 77.42\%& 83.29\%& 70.26\%& 74.09\%\\
\method-H/\plugin{MLA} & 74.25\%&68.12\%&\second{84.40\%}&83.57\%& 77.82\% & 83.64\% &\second{73.67\%} & \second{78.24}\%  \\
\method/\plugin{MLA}   & \first{74.93\%} & \second{68.90\%} & \first{84.46\%} & \first{83.75\%} & \first{82.34\%} & \first{86.64\%} & \first{74.87\%} & \first{80.06\%} \\
\Xcline{1-9}{1pt}
\end{tabular}
\end{table*}

\subsection{Experimental Settings}
\subsubsection{Baselines}
To demonstrate the superiority of our proposed method, we choose a wide range of methods as baselines for experiments. They are OGR-GB~\cite{OGR-GB:conf/cvpr/WangTF20}, OGM~\cite{OGM:conf/cvpr/PengWD0H22}, DOMFN~\cite{DOMFN:conf/mm/0074ZGGZ22}, MSES~\cite{MSES:conf/acpr/FujimoriEKM19}, PMR~\cite{PMR:conf/cvpr/Fan0WW023}, AGM~\cite{AGM:conf/iccv/LiLHLLZ23}, MSLR~\cite{MSLR:conf/acl/YaoM22}, ReconBoost~\cite{ReconBoost:conf/icml/CongHua24}, SMV~\cite{SMV:conf/cvpr/YakeRZD24}, DI-MML~\cite{DI-MML:conf/mm/FanXWLG24} and MLA~\cite{MLA:conf/cvpr/ZhangYBY24}. Among these methods, all baselines except SMV are the solutions from the model learning perspective. 


\subsubsection{Evaluation Protocols}
We adopt accuracy~(Acc.) and macro-F1~(MacF1) as metrics for \Twitter, \Sarcasm, and \NVGesture~datasets following the setting of the paper~\cite{Sarcasm:conf/acl/CaiCW19,Twitter15:conf/ijcai/Yu019}. Furthermore, we select accuracy, mean average precision~(MAP) as evaluation metrics for \CREMAD~and \Kinetics~datasets following the setting of OGM~\cite{OGM:conf/cvpr/PengWD0H22}. The accuracy is used to measure the proportion of ground-truth labels that the model predicts correctly. MAP is calculated by taking the mean of average precision. And the MacF1 can be calculated by averaging the F1 score for each class.

\subsubsection{Implementation Details} Following the setting of~\cite{Twitter15:conf/ijcai/Yu019,Sarcasm:conf/acl/CaiCW19}, we employ ResNet50~\cite{ResNet:conf/cvpr/HeZRS16} as the image feature extractor and BERT~\cite{BERT:conf/naacl/DevlinCLT19} as text feature extractor on the dataset \Twitter~and \Sarcasm~datasets for image and text modalities. Furthermore, we also adopt image and text encoder from pretrained models CLIP~\cite{CLIP:conf/icml/RadfordKHRGASAM21} to verify the effectiveness of the large-scale pretrained vision-language model. Following the setting of OGM, we use ResNet18~\cite{ResNet:conf/cvpr/HeZRS16} as the feature extractor to encode audio and video for \CREMAD~and \Kinetics~datasets. For the last three modalities dataset \NVGesture, we continue the setting of the previous paper~\cite{nvGesture:conf/icml/WuJCG22} and take the I3D~\cite{I3D:conf/cvpr/CarreiraZ17} as unimodal feature extractor. For a fair comparison, all methods use the same feature extractor for the experiment. And the classification head is only composed of one linear layer after the feature extractor. The hidden dimension of features about the audio and video is 512 when the text, image, and \NVGesture~is 1024. For \CREMAD, \Kinetics~and \NVGesture~datasets, we adopt the SGD optimizer with the momentum of $0.9$ and weight decay of $1\times 10^{-4}$. At the beginning, the learning rate is $1\times 10^{-2}$ and will be divided by 10 when the loss is saturated. For \Twitter~and \Sarcasm~datasets, we use Adam as the optimizer and set the learning rate as $1\times 10^{-5}$. The learning rate of RL models is always set to $1\times 10^{-4}$ through using the cross-validation strategy with a validation set. $\beta$~and $~\alpha$ is always set to 0.5 and $\frac{1}{T} \times \ln(\frac{1}{\beta} ) $ .The batch size is set to 64, except for the \NVGesture~dataset which is set to 6 due to the out-of-memory issue. All experiments are performed with an NVIDIA RTX 3090 GPU.

\subsection{Main Results}
{\noindent\bf Comparison with SOTA MML Baselines:} To substantiate the superiority of \method, we conduct comprehensive comparisons with diverse baselines, including unimodal methods, Baseline, and multimodal learning methods with rebalanced strategy. The Baseline denotes the vanilla multimodal learning approach which minimizes the unimodal and multimodal losses. For \method, we integrate our method with the baseline and a competitive baseline MLA. Specifically, we denote the RL-based \method~with the baseline and MLA as ``\method/\plugin{Baseline}'' and ``\method/\plugin{MLA}'', respectively. The heuristic \method~with MLA is denoted as ``\method-H/\plugin{MLA}''.


The results for the first four datasets are presented in Table~\ref{tab:main-exp}. In Table~\ref{tab:main-exp}, the unimodal results are based on text and image modality for \Twitter~and \Sarcasm~datasets. For \CREMAD~and \Kinetics~datasets, the unimodal results are based on video and audio modalities. From Table~\ref{tab:main-exp}, we can draw the following observations: (1). By comparing multimodal learning methods with unimodal methods, we observe the superiority of the former over the latter in almost all cases. However, due to modality imbalance, unimodal methods occasionally outperform MML methods, which is highlighted by the results with a gray background; (2). Our heuristic \method-H/\plugin{MLA} can outperform MLA in most cases. However, on \CREMAD~dataset, the performance of \method-H/{\small MLA} is worse than that of MLA. One possible reason is that the heuristic strategy fails to capture the dynamic change in the amount of sampled data for this dataset. We further explore this issue in Section~\ref{sec:further-analysis}; (3). \method/\plugin{Baseline} can outperform Baseline in all cases and achieve competitive performance compared with MML approaches with rebalanced strategy, demonstrating the importance of the data sampling; (4). By integrating with MLA, \method/\plugin{MLA} can achieve the best performance in almost all cases compared with all baselines, demonstrating the effectiveness of our proposed method. 

In Table~\ref{tab:three-modals}, we report the accuracy and MacF1 on \NVGesture~dataset with three modalities. From Table~\ref{tab:three-modals}, we can see that our \method/\plugin{MLA}~can seamlessly extend to the scenario with multiple modalities and achieve the best performance in most cases. 

\begin{table}[t]
\centering
\caption{Comparison with SOTA MML baselines on \NVGesture~dataset. The results are marked similarly to those in Table~\ref{tab:main-exp}.} 
\label{tab:three-modals}
\begin{tabular}{ll|cc}
\Xcline{1-4}{1pt}\hline
\multicolumn{2}{c|}{{Method}} & Acc.  & MacF1\\
\hline\hline
&RGB     &78.22\% & 78.33\%  \\
Unimodal&OF     & 78.63\% & 78.65\%  \\
&Depth     &  81.54\% & 81.83\%  \\
\hline\multirow{8}{*}{Multimodal}
&OGR-GB~\cite{OGR-GB:conf/cvpr/WangTF20}  & 82.99\% & 83.05\%  \\
&MSES~\cite{MSES:conf/acpr/FujimoriEKM19}      & \unimodal{81.12\%} & \unimodal{81.47\%}  \\
&AGM~\cite{AGM:conf/iccv/LiLHLLZ23}       & 82.78\% & 82.82\%  \\
&MSLR~\cite{MSLR:conf/acl/YaoM22}      & 82.86\% & 82.92\%  \\
&SMV~\cite{SMV:conf/cvpr/YakeRZD24}       & 83.52\% & 83.41\%\\
&ReconBoost~\cite{ReconBoost:conf/icml/CongHua24}& \second{84.13\%} &\first{86.32\%}\\
&MLA~\cite{MLA:conf/cvpr/ZhangYBY24}       & 83.73\% & 83.87\%  \\
\cline{2-4}
&\method/\plugin{MLA}  & \first{84.25\%} & \second{85.36\%}  \\
\Xcline{1-4}{1pt}
\end{tabular}
\end{table}

\begin{table}[t]
\centering
\caption{Algorithm adaptability on \Kinetics~dataset.} 
\label{tab:adaptability}
\begin{tabular}{l|ll}
\Xcline{1-3}{1pt}\hline
Method                  & Acc.   & MAP  \\
\hline\hline
PMR                     & 66.56\%& 71.93\%\\
\method/\plugin{PMR}    & \first{70.13\%}/{\gain{3.57\%}}& \first{74.36\%}/{\gain{2.43\%}}        \\\hline
AGM                     & 66.02\%& 72.52\%\\
\method/\plugin{AGM}    & \first{69.12\%}/{\gain{3.10\%}}& \first{74.97\%}/{\gain{2.45\%}}       \\\hline
DI-MML                  & 72.03\%& 76.24\%\\
\method/\plugin{DI-MML} & \first{74.15\%}/{\gain{2.12\%}} & \first{78.22\%}/{\gain{1.98\%}} \\\hline
\Xcline{1-3}{1pt}
\end{tabular}
\end{table}

{\noindent\bf Algorithm Adaptability:} We further integrate our method with other two representative multimodal learning approaches including PMR~\cite{PMR:conf/cvpr/Fan0WW023}, AGM~\cite{AGM:conf/iccv/LiLHLLZ23}, and DI-MML~\cite{DI-MML:conf/mm/FanXWLG24}, to verify the algorithm adaptability. PMR and AGM focus on leveraging the gradient to facilitate multimodal learning. And Similar to MLA, DI-MML is a multimodal learning method built on an alternating optimization paradigm. The performance and the improvement on \Kinetics~dataset are presented in Table~\ref{tab:adaptability}. The results in Table~\ref{tab:adaptability} demonstrate that our method can further improve the performance and exhibits strong adaptability.

\subsection{Ablation Study}
To fully exploit the effectiveness of \method, we analyze the impact of two important components, i.e., reinforcement learning and cumulative modality discrepancy score. The results are shown in Table~\ref{tab:ablation} on \Kinetics~dataset, where we utilize baseline to denote the multimodal learning without reinforcement learning and discrepancy score~(DS), {and ``\method~w/o RL'' to denote the approach which directly uses discrepancy score to calculate the percentage of batch size for each modalities\footnote{Since the cumulative discrepancy score is the input of reinforcement learning, the method with reinforcement learning but without discrepancy score cannot be performed.}. From Table~\ref{tab:ablation}, we can find that directly adopting discrepancy score to guide the unimodal sampling can improve the overall performance by comparing \method~w/o RL with baseline. Furthermore, by comparing \method~with \method~w/o RL, we can find that the overall performance is further improved by using reinforcement learning-based adaptive unimodal sampling. The results on the rest datasets and other details are provided in the appendix.

\begin{table}[t]
\centering
\caption{Performance comparison on \Kinetics~dataset for ablation study.}  
\label{tab:ablation}
\begin{tabular}{L{1.25cm}|L{1.85cm}|C{0.275cm}C{0.275cm}|C{0.9cm}C{0.9cm}}
\Xcline{1-6}{1pt}
Modality&Method&DS&RL&Acc. & MAP \\
\hline\hline
&Baseline      &\ding{56} & \ding{56} & 55.72\%&54.23\%  \\
 Audio&\method~w/o RL&\ding{52} & \ding{56} & 56.12\%&57.37\%  \\   
&\method       &\ding{52} & \ding{52} & \bf{57.05\%}&\bf{61.77\%} \\\hline
&Baseline      &\ding{56} & \ding{56} & 54.02\%&56.35\%\\
Video&\method~w/o RL&\ding{52} & \ding{56} & 54.17\%&57.39\%\\
&\method       &\ding{52} & \ding{52} & \bf{55.00\%}&\bf{57.95\%} \\\hline
&Baseline      &\ding{56} & \ding{56} & 70.04\%&74.13\% \\
Multi&\method~w/o RL&\ding{52} & \ding{56} & 73.44\%&77.15\% \\
&\method       &\ding{52} & \ding{52} & \bf{74.87\%}&\bf{80.06\%}\\
\Xcline{1-6}{1pt}
\end{tabular}
\end{table}

\begin{figure}[t] 
\centering
\begin{minipage}{.4875\linewidth}
\centering
\includegraphics[width=\linewidth]{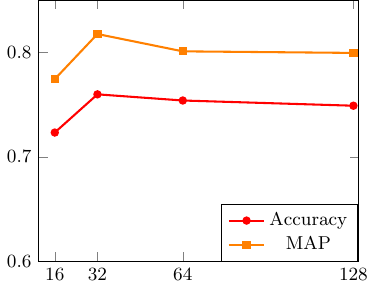}\\
{(a). Constant batch size $N_B$.}
\end{minipage} 
\begin{minipage}{.48\linewidth}
\centering
\includegraphics[width=0.985\linewidth]{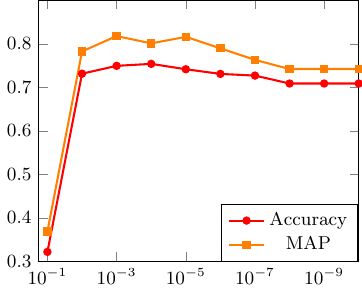}\\
{(b). Learning rate.}
\end{minipage}
\caption{Impact of constant batch size $N_B$ and learning rate.}
\label{fig:sensitivity}
\end{figure}

\begin{figure*}[t] 
\centering
\begin{minipage}{.245\linewidth}
\centering
\includegraphics[width=\linewidth]{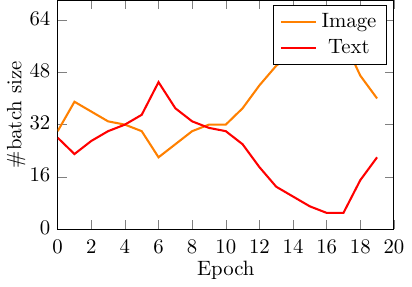}\\
{(a). \Twitter.}
\end{minipage} 
\begin{minipage}{.245\linewidth}
\centering
\includegraphics[width=\linewidth]{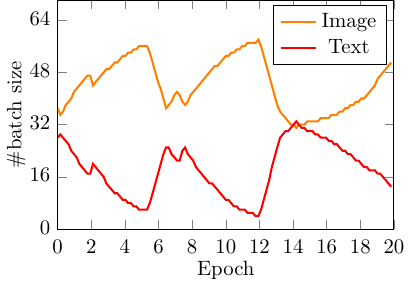}\\
{(b). \Sarcasm.}
\end{minipage}
\begin{minipage}{.245\linewidth}
\centering
\includegraphics[width=\linewidth]{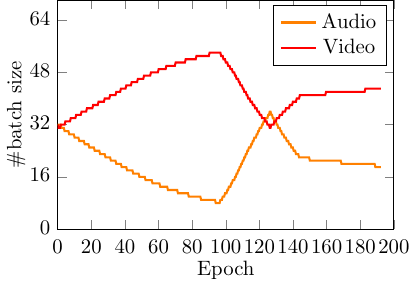}\\
{(c). \CREMAD.}
\end{minipage}
\begin{minipage}{.245\linewidth}
\centering
\includegraphics[width=\linewidth]{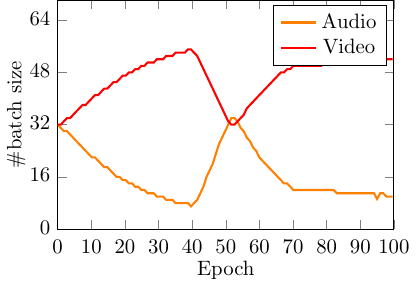}\\
{(d). \Kinetics.}
\end{minipage}
\caption{Change of batch size during the training process. Best viewed in color.}
\label{fig:change-of-batch-size}
\end{figure*}

\begin{figure*}[t] 
\centering
\begin{minipage}{.245\linewidth}
\centering
\includegraphics[width=\linewidth]{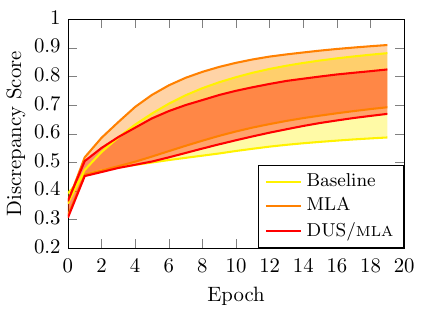}\\
{(a). \Twitter.}
\end{minipage} 
\begin{minipage}{.245\linewidth}
\centering
\includegraphics[width=\linewidth]{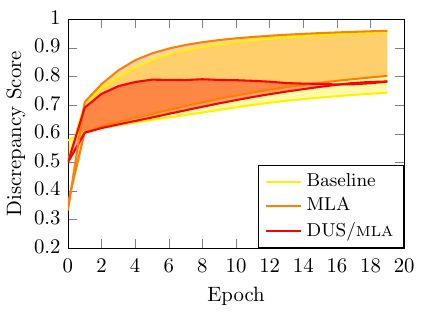}\\
{(b). \Sarcasm.}
\end{minipage}
\begin{minipage}{.245\linewidth}
\centering
\includegraphics[width=\linewidth]{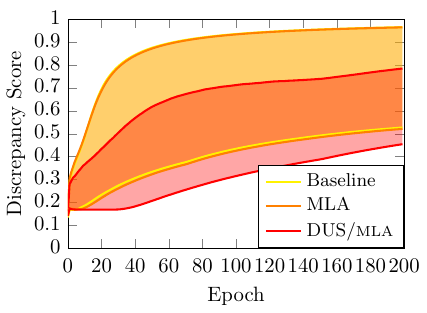}\\
{(c). \CREMAD.}
\end{minipage}
\begin{minipage}{.245\linewidth}
\centering
\includegraphics[width=\linewidth]{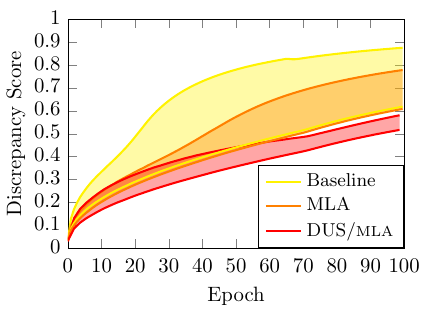}\\
{(d). \Kinetics.}
\end{minipage}
\caption{Change of cumulative modality discrepancy score during the training process. Best viewed in color.}
\label{fig:change-of-discrepancy-score}
\end{figure*}

\subsection{Sensitivity to Hyper-Parameters}
In this section, we explore the impact of hyper-parameter constant batch size $N_B$ and the learning rate of the reinforcement learning procedure. 

According to our design, we use a constant $N_B$ to denote the total data points for all modalities. We explore the performance with different constant batch size $N_B$. In Figure~\ref{fig:sensitivity}~(a), we present the accuracy and MAP on \Kinetics~dataset with different $N_B\in\{16,32,64,128\}$. We can find that \method~can achieve close accuracy/MAP and is not sensitive to constant batch size when $N_B>32$.

Furthermore, since reinforcement learning can be treated as a plugin for \method, we present the impact of the most important parameter learning rate on \Kinetics~dataset. We report the results in Figure~\ref{fig:sensitivity}~(b) with different learning rate in $\{10^{-r}\}_{r=1}^{10}$. From Figure~\ref{fig:sensitivity}~(b), we can see that our method is not sensitive to the learning rate when the learning rate is smaller than $10^{-1}$.
\subsection{Further Analysis}\label{sec:further-analysis}
{\noindent\bf Change of Batch Size:} We visualize the change of the quantity of data sampled for all modalities during the learning process to reveal the pattern of information variation during model training. Specifically, we illustrate the change of batch size on \CREMAD, \Kinetics, \Twitter, and \Sarcasm~datasets in Figure~\ref{fig:change-of-batch-size}. From Figure~\ref{fig:change-of-batch-size}, we can observe that: (1). During the training process, the batch size guided by the discrepancy metric is dynamically changed. (2). The change in batch size does not show a monotonically increasing or decreasing trend. This may explain why our heuristic method did not achieve the best results in some cases. (3). Interestingly, at certain stages of training~(around the 4th and 8th epoch) on \Twitter~dataset, the batch size of the text modality is larger than that of the image modality, contrary to the overall trend. One possible reason is that at this stage, the confidence amplitude of the image modality increases, leading to fewer samples being needed to balance the learning.

{\noindent\bf Change of Discrepancy Score: }To fully explore the modality imbalance phenomenon during the training process, we visualize the change of cumulative discrepancy score during training on \CREMAD, \Kinetics, \Twitter, and \Sarcasm~datasets for baseline, MLA, and \method/\plugin{MLA}. The results are provided in Figure~\ref{fig:change-of-discrepancy-score}, where the solid lines of the same color are used to represent the discrepancy scores of the two modalities, i.e., audio/video for \CREMAD/\Kinetics~datasets, and image/text for \Twitter/\Sarcasm~datasets. The light shaded areas are used to indicate the gap in cumulative modality discrepancy scores. From Figure~\ref{fig:change-of-discrepancy-score}, we can draw the following observations: (1). The discrepancy score shows an overall upward trend, as it indirectly reflects the model's prediction ability. (2). The discrepancy score gap of \method/\plugin{MLA}~is smaller than that of MLA and baseline. This is likely due to our adjustment of the data amount during the modality learning process, leading to a more balanced learning process.



\section{Conclusion}\label{sec:conclusion}
In this paper, we propose a novel multimodal learning approach, called data-aware unimodal sampling~(\method). By designing a cumulative discrepancy score that averages the model's predictions of the ground-truth class, we can monitor the learning process in multimodal learning during training. Based on the discrepancy score, we propose a heuristic and a reinforcement learning-based balanced data-aware unimodal sampling approach. To this end, we further alleviate the modality imbalance problem from the data sampling perspective for multimodal learning, thus leading to better performance. Our \method~can be seamlessly integrated with almost all existing MML methods as a plugin. Extensive experiments on widely used datasets show that our proposed \method~can achieve the best performance by comparing with various SOTA baselines.

{
    \small
    \bibliographystyle{ieeenat_fullname}
    \bibliography{main}
}
\clearpage
\begin{center}
    \Large
    Appendix of Paper
    \\[10pt]
    \normalsize
\end{center}
\setcounter{figure}{0}
\setcounter{table}{0}
\setcounter{section}{0}
\renewcommand*{\thesection}{\Alph{section}}
\renewcommand{\thefigure}{A\arabic{figure}}
\renewcommand{\thetable}{A\arabic{table}}

\section{Additional Experimental Results}
We report more detailed experimental results in this section, including main results with $Mean$ and $Std$, the impact of cumulative strategy, and ablation study on \CREMAD, \Twitter, and \Sarcasm~datasets.
\subsection{Main Results with $Mean$ and $Std$}

We report report the average performance along with the standard deviation in Table~\ref{tab:main-exp} to eliminate the effects of randomness. Concretely, all experiments are run three times and average performance is reported on all datasets. In Table~\ref{tab:main-exp}, we report accuracy~(Acc.) as well as Macro-F1 for \Twitter~and \Sarcasm~datasets, and the accuracy as well as MAP for \CREMAD, \Kinetics~and \NVGesture~datasets. From the experimental results we can see that our method exhibits robust performance. 

\begin{table}[!htb]
\caption{Detailed performance with $Mean$ and $Std$ values on all datasets.}
\label{tab:main-exp-2}
\centering
\begin{tabular}{l|c|c} 
\Xcline{1-3}{1pt}
\multirow{2}{*}{Dataset} & \multicolumn{2}{c}{\method}  \\
\cline{2-3}
                          & Acc.                 & Macro-F1/MAP               \\
\hline\hline
\Twitter      & 74.93\%$\pm$1.30\% &  68.90\%$\pm$1.52\% \\\hline
\Sarcasm                  & 84.46\%$\pm$0.93\% &   83.75\%$\pm$0.69\% \\\hline
\CREMAD                   & 82.34\%$\pm$0.92\% & 86.64\%$\pm$1.15\% \\\hline
\Kinetics                 & 74.87\%$\pm$0.82\% &       80.06\%$\pm$0.61\% \\\hline
\NVGesture                & 84.25\%$\pm$0.52\% &     85.36\%$\pm$0.77\% \\
\Xcline{1-3}{1pt}
\end{tabular}
\end{table}

\subsection{Impact of Cumulative Strategy}
Taking into account the randomness of each single batch, we design a cumulative OGM score $\hat s_t^{(j)}$ in the paper. We exploit the effectiveness of this cumulative OGM score. Specifically, we conduct an experiment to compare the method with and without cumulative strategy. These two methods are denoted as \method~and \method~w/o CS. And the results on \Kinetics~dataset are reported in Table~\ref{tab:cs}. From Table~\ref{tab:cs}, we can find that \method~can achieve better performance compared to \method~w/o CS, demonstrating the effectiveness of cumulative strategy.

\begin{table}[!htb]
\centering
\caption{Impact of cumulative strategy on \Kinetics~dataset.}  
\label{tab:cs}
\begin{tabular}{c|l|cc}
\Xcline{1-4}{1pt}
 Modal&Method&Acc. & MAP \\
\hline\hline
\multirow{2}{*}{Audio}
&\method~w/o CS&  56.81\%&60.45\%  \\   
&\method    & \bf{57.05\%}&\bf{61.77\%}  \\\hline
\multirow{2}{*}{Video}
&\method~w/o CS& 54.72\%&57.69\%\\
&\method    & \bf{55.00\%}&\bf{57.95\%}\\\hline
\multirow{2}{*}{Multimodal}
&\method~w/o CS& {74.53\%}&79.42\% \\
&\method    & \bf{74.87\%}&\bf{80.06\%}\\
\Xcline{1-4}{1pt}
\end{tabular}
\end{table}

\subsection{Ablation Study}
To further study the effectiveness of our method, we report the ablation study results on all datasets except \Kinetics. Specifically, the results on \CREMAD~dataset and \Twitter~as well as \Sarcasm~datasets are presented in Table~\ref{tab:ablation-cremad} and Table~\ref{tab:ablation-twitter-sarcasm}, respectively. For \CREMAD~dataset, we report the accuracy~(Acc.) and MAP following the setting of OGM~\cite{OGM:conf/cvpr/PengWD0H22}. And for \Twitter~and \Sarcasm~datasets, we report the accuracy and Macro-F1. The results demonstrate the effectiveness of key compoenents of our method in almost all cases. 

\begin{table}[!htbp]
\centering
\caption{Performance comparison on \CREMAD~dataset for ablation study.}  
\label{tab:ablation-cremad}
\begin{tabular}{L{1.25cm}|L{1.85cm}|C{0.275cm}C{0.275cm}|C{0.9cm}C{0.9cm}}
\Xcline{1-6}{1pt}
Modality&Method&DS&RL&Acc. & MAP \\
\hline\hline
&Baseline      &\ding{56} & \ding{56} & 57.27\%&61.56\%  \\
 Audio&\method~w/o RL&\ding{52} & \ding{56} & 60.21\%&63.58\% \\   
&\method       &\ding{52} & \ding{52} & \bf{60.88\%}&\bf{65.27\%} \\\hline
  &Baseline      &\ding{56} & \ding{56} & 64.91\%&69.34\% \\
Video&\method~w/o RL&\ding{52} & \ding{56} & 68.54\%&77.91\% \\
&\method       &\ding{52} & \ding{52} & \bf{72.44\%}&\bf{80.77\%} \\\hline
&Baseline      &\ding{56} & \ding{56} & 79.43\%&85.72\% \\
Multi&\method~w/o RL&\ding{52} & \ding{56} & 80.64\%&\bf{87.21\%} \\
&\method       &\ding{52} & \ding{52} & \bf{82.34\%}&{86.64\%}\\
\Xcline{1-6}{1pt}
\end{tabular}
\end{table}

\begin{table*}[!htb]
\centering
\caption{Accuracy and Macro-F1 for ablation study.}  
\label{tab:ablation-twitter-sarcasm}
\begin{tabular}{c|l|cc|cc|cc|cc}
\Xcline{1-10}{1pt}
 \multirow{2}{*}{Dataset}&\multirow{2}{*}{Method}&\multirow{2}{*}{DS}  &\multirow{2}{*}{RL} & \multicolumn{2}{c|}{Text} & \multicolumn{2}{c|}{Image} & \multicolumn{2}{c}{Multimodal}\\\cline{5-10}
&&&&Acc. & Macro-F1 &Acc. & Macro-F1 &Acc. & Macro-F1 \\
\hline\hline
\multirow{3}{*}{\Twitter}
&Baseline      &\ding{56} & \ding{56} & 73.84\%&68.61\% & 58.89\%&43.87\% & 73.52\%&67.13\% \\
&\method~w/o RL&\ding{52} & \ding{56} & 73.90\%&69.02\% & 59.77\%&44.29\% & 74.12\%&67.23\% \\   
&\method       &\ding{52} & \ding{52} & \bf{74.12\%}&\bf{69.58\%} & \bf{60.43\%}&\bf{44.87\%} & \bf{74.93\%}&\bf{68.90\%}\\\hline
\multirow{3}{*}{\Sarcasm}
&Baseline      &\ding{56} & \ding{56} & 81.72\%&80.79\% & 72.23\%&71.14\% & 84.26\%&83.48\% \\
&\method~w/o RL&\ding{52} & \ding{56} & 82.39\%&81.52\% & 72.50\%&72.43\% & 84.32\%&83.57\% \\   
&\method       &\ding{52} & \ding{52} & \bf{83.06\%}&\bf{82.57\%} & \bf{73.59\%}&\bf{73.26\%} & \bf{84.46\%}&\bf{83.75\%}\\
\Xcline{1-10}{1pt}
\end{tabular}
\end{table*}

\section{Limitations}
{\noindent\bf Conditions for Integrating with the MML Method: }Since for our algorithm, the quantities of samples for different modalities in each batch are inconsistent. Consequently, if the model learning-based MML methods rely exclusively on paired multimodal data as input, they cannot be directly integrated with our approach. For the method which utilizes both paired data-based loss and unpaired data-based loss, such as DI-MML~\cite{DI-MML:conf/mm/FanXWLG24}, we can split the training data in each batch as paired data and unpaired data as input of the loss.
\end{document}